\documentclass[11pt,a4paper]{article}
\usepackage[hyperref]{acl2018}
\usepackage{times}
\usepackage{latexsym}

\usepackage{url}
\usepackage{graphicx}
\usepackage{amsmath}
\usepackage{multirow}
\usepackage{booktabs}  
\usepackage{graphicx}
\usepackage{algorithm}  
\usepackage{algpseudocode} 
\usepackage{subfigure}
\usepackage{natbib}
\usepackage{pgfplots}
\usepackage{pgfplotstable}
\pgfplotsset{width=7.0cm,compat=1.5}
\usepackage{color,xcolor,framed}
\usepackage{bm}
\usepackage{CJK}
\newcommand*{\affaddr}[1]{#1} 
\newcommand*{\affmark}[1][*]{\textsuperscript{#1}}
\newcommand*{\email}[1]{\texttt{#1}}

\aclfinalcopy 
\def\aclpaperid{1235} 

\newcommand{\cev}[1]{\reflectbox{\ensuremath{\vec{\reflectbox{\ensuremath{#1}}}}}}

\title{Bag-of-Words as Target for Neural Machine Translation}

\author{Shuming Ma\affmark[1], Xu Sun\affmark[1,2], Yizhong Wang\affmark[1], Junyang Lin\affmark[3]\\
\affaddr{\affmark[1]MOE Key Lab of Computational Linguistics, School of EECS, Peking University}\\
\affaddr{\affmark[2]Deep Learning Lab, Beijing Institute of Big Data Research, Peking University}\\
\affaddr{\affmark[3]School of Foreign Languages, Peking University}\\
\email{\{shumingma, xusun, yizhong, linjunyang\}@pku.edu.cn}\\
}

\date{}

\begin{document}
\maketitle
\begin{abstract}

A sentence can be translated into more than one correct sentences. 
However, most of the existing neural machine translation models only use one of the correct translations as the targets, and the other correct sentences are punished as the incorrect sentences in the training stage. 
Since most of the correct translations for one sentence share the similar bag-of-words, it is possible to distinguish the correct translations from the incorrect ones by the bag-of-words. 
In this paper, we propose an approach that uses both the sentences and the bag-of-words as targets in the training stage, in order to encourage the model to generate the potentially correct sentences that are not appeared in the training set. 
We evaluate our model on a Chinese-English translation dataset, and experiments show our model outperforms the strong baselines by the BLEU score of 4.55.\footnote{The code is available at \url{https://github.com/lancopku/bag-of-words}}

\end{abstract}

\section{Introduction}

Neural Machine Translation (NMT) has achieve success in generating coherent and reasonable translations. Most of the existing neural machine translation systems are based on the sequence-to-sequence model~\citep{seq2seq}. The sequence-to-sequence model (Seq2Seq) regards the translation problem as the mapping from the source sequences to the target sequences. The encoder of Seq2Seq compresses the source sentences into the latent representation, and the decoder of Seq2Seq generates the target sentences from the source representations. The cross-entropy loss, which measures the distance of the generated distribution and the target distribution, is minimized in the training stage, so that the generated sentences are as similar as the target sentences.

\begin{CJK}{UTF8}{gbsn}
\begin{table}[t]
\centering
    \begin{tabular}{p{7.2cm}}
    \hline
    \textbf{Source:}  今年前两月广东高新技术产品出口37.6亿美元。\\
    \hline
    \textbf{Reference:} Export of high - tech products in guangdong in first two months this year reached 3.76 billion us dollars .\\
    \hline
    \textbf{Translation 1:} Guangdong 's export of new high technology products amounts to us \$3.76 billion in first two months of this year .\\
    \hline
    \textbf{Translation 2:} Export of high - tech products has frequently been in the spotlight , making a significant contribution to the growth of foreign trade in guangdong .\\
    \hline
    \end{tabular}
    \caption{An example of two generated translations. Although Translation 1 is much more reasonable, it is punished more severely than Translation 2 by Seq2Seq.}
    \label{intro_example}
\end{table}
\end{CJK}

Due to the limitation of the training set, most of the existing neural machine translation models only have one reference sentences as the targets. However, a sentence can be translated into more than one correct sentences, which have different syntax structures and expressions but share the same meaning. 
The correct translations that are not appeared in the training set will be punished as the incorrect translation by Seq2Seq, which is a potential harm to the model. 
Table~\ref{intro_example} shows an example of two generated translations from Chinese to English. Translation 1 is apparently more proper as the translation of the source sentence than Translation 2, but it is punished even more severely than Translation 2 by Seq2Seq.

Because most of the correct translations for one source sentence share the similar bag-of-words, it is possible to distinguish the correct translations from the incorrect ones by the bag-of-words in most cases. In this paper, we propose an approach that uses both sentences and bag-of-words as the targets. In this way, the generated sentences which cover more words in the bag-of-words (e.g. Translation 1 in Table~\ref{intro_example}) are encouraged, while the incorrect sentences (e.g. Translation 2) are punished more severely. We perform experiments on a popular Chinese-English translation dataset. Experiments show our model outperforms the strong baselines by the BLEU score of 4.55.

\section{Bag-of-Words as Target}

In this section, we describe the proposed approach in detail.

\subsection{Notation}

Given a translation dataset that consists of $N$ data samples, the $i$-th data sample ($x^{i}$, $y^{i}$) contains a source sentence $x^{i}$, and a target sentence $y^{i}$. The bag-of-words of $y^{i}$ is denoted as $b^{i}$. The source sentence $x^{i}$, the target sentence $y^{i}$, and the bag-of-words $b^{i}$ are all sequences of words:
\begin{equation*}
x^{i}=\{x^{i}_{1},x^{i}_{2},...,x^{i}_{L_{i}}\}
\end{equation*}
\begin{equation*}
y^{i}=\{y^{i}_{1},y^{i}_{2},...,y^{i}_{M_{i}}\}
\end{equation*}
\begin{equation*}
b^{i}=\{b^{i}_{1},b^{i}_{2},...,b^{i}_{K_{i}}\}
\end{equation*}
where $L_{i}$, $M_{i}$, and $K_{i}$ denote the number of words in $x^{i}$, $y^{i}$, and $b^{i}$, respectively.

The target of our model is to generate both the target sequence $y^{i}$ and the corresponding bag-of-words $b^{i}$. For the purpose of simplicity, $(\bm{x}, \bm{y}, \bm{b})$ is used to denote each data pair in the rest of this section.

\begin{figure}[tb]
	\centering
    \includegraphics[width=1.0\linewidth]{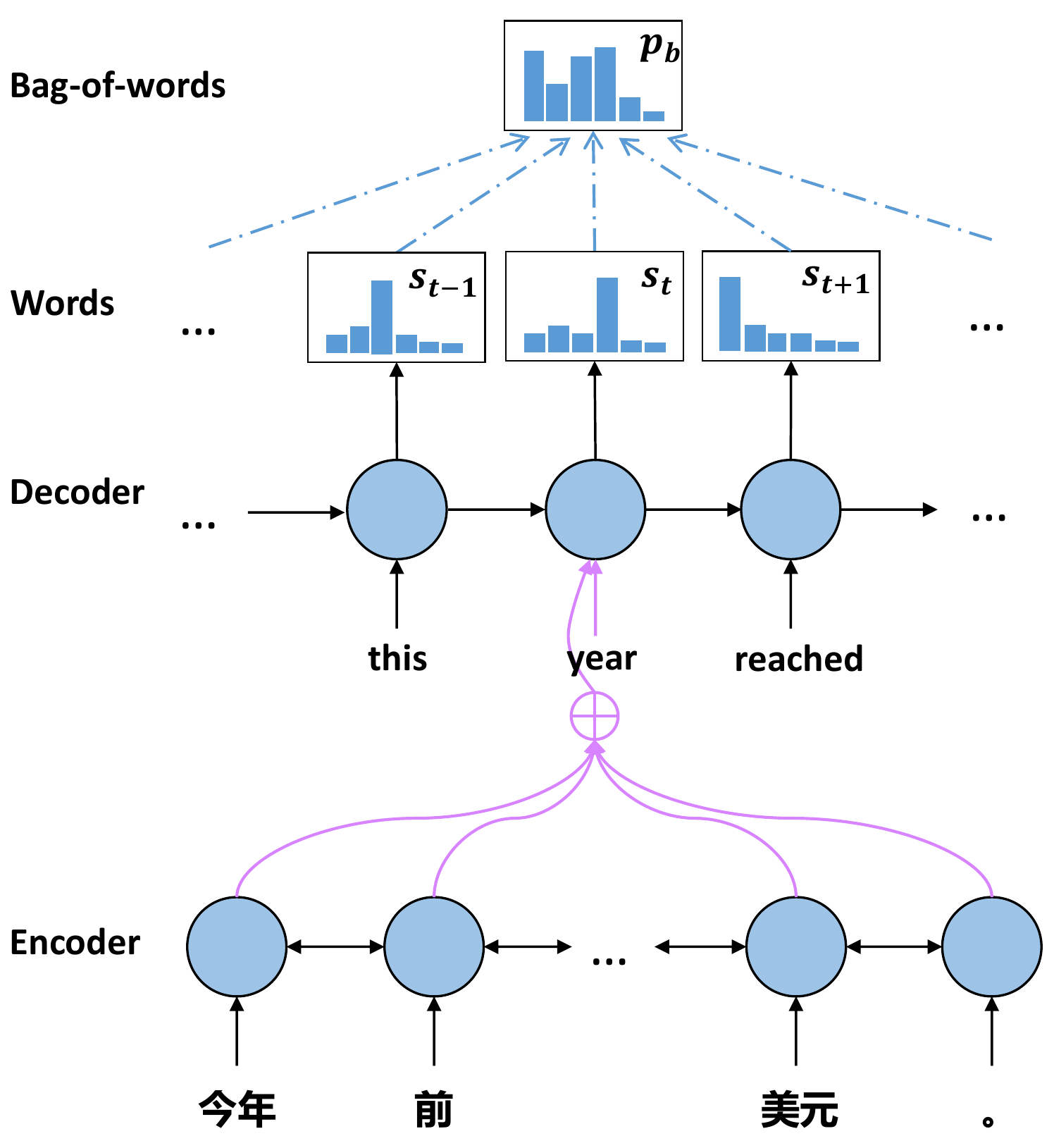}
	\caption{The overview of our model. The encoder inputs the source sentence, and the decoder outputs the word distribution at each position. The distribution of all position is summed up to a sentence-level score, which can be used to generate the bag-of-words.}\label{model_fig}
\end{figure}

\subsection{Bag-of-Words Generation}

We regard the bag-of-words generation as the multi-label classification problem. We first perform the encoding and decoding to obtain the scores of words at each position of the generated sentence. Then, we sum the scores of all positions as the sentence-level score. Finally, the sentence-level score is used for multi-label classification, which identifies whether the word appears in the translation.

In our model, the encoder is a bi-directional Long Short-term Memory Network (BiLSTM), which produces the representation $\bm{h}=\{h_1,h_2,...,h_L\}$ from the source text $\bm{x}$:
\begin{equation}\label{flstm}
\vec{h}_t = \vec{f}(x_t,\vec{h}_{t-1})
\end{equation}
\begin{equation}\label{blstm}
\cev{h}_t = \cev{f}(x_t,\cev{h}_{t+1})
\end{equation}
\begin{equation}\label{clstm}
h_t = \vec{h}_t+\cev{h}_t
\end{equation}
where $\vec{f}$ and $\cev{f}$ are the forward and the backward functions of LSTM for one time step, $\vec{h}_t$ and $\cev{h}_t$ are the forward and the backward hidden outputs respectively, $x_t$ is the input at the $t$-th time step, and $L$ is the number of words in sequence $\bm{x}$.

The decoder consists of a uni-directional LSTM, with an attention, and a word generator. The LSTM generates the hidden output $q_t$:
\begin{equation}\label{flstm}
q_t = f(y_{t-1},q_{t-1})
\end{equation}
where $f$ is the function of LSTM for one time step, and $y_{t-1}$ is the last generated words at $t$-th time step.
The attention mechanism~\citep{luong2015stanford} is used to capture the source information:
\begin{equation}\label{attention1}
v_{t}=\sum_{i=1}^{N}{\alpha_{ti}h_{i}}
\end{equation}
\begin{equation}\label{attention2}
\alpha_{ti}=\frac{e^{g(q_{t},h_{i})}}{\sum_{j=1}^{N}{e^{g(q_{t},h_{j})}}}
\end{equation}
\begin{equation}\label{attention3}
g(q_{t},h_{i})=\tanh{(q^{T}_{t}W_{t}h_{i})}
\end{equation}
where $W_{t}$ is a trainable parameter matrix.
Then, the word generator is used to compute the probability of each output word at $t$-th time step:
\begin{equation}
p_{w_{t}}=softmax(s_t)
\end{equation}
\begin{equation}
s_t=W_{g}v_{t}+b_{g}
\end{equation}
where $W_{g}$ and $b_{g}$ are parameters of the generator.

To get a sentence-level score for the generated sentence, we generate a sequence of word-level score vectors $s_t$ at all positions with the output layer of decoder, and then we sum up the word-level score vectors to obtain a sentence-level score vector. Each value in the vector represents the sentence-level score of the corresponding word, and the index of the value is the index of the word in the dictionary. After normalizing the sentence-level score with sigmoid function, we get the probability for each word, which represents how possible the word appears in the generated sentence regardless of the position in the sentence. Compared with the word-level probability $p_{w_{t}}$, the sentence-level probability $p_{b}$ of each word is independent of the position in the sentence.

More specifically, the sentence-level probability of the generated bag-of-words $p_{b}$ can be written as:
\begin{equation}\label{sigmoid}
p_{b}=sigmoid(\sum_{t=1}^{M}s_t)
\end{equation}
where $M$ is the number of words in the target sentence.


\subsection{Targets and Loss Function}

We have two targets at the training stage: the reference translation (appears in the training set) and the bag-of-words.
The bag-of-words is used as the approximate representation of the correct translations that do not appear in the training set. 
For the targets, we have two parts of loss functions:
\begin{equation}
l_{1}=-\sum_{t=1}^{M}y_{t}\log{p_{w_{t}}(y_{t})}
\end{equation}
\begin{equation}
l_{2}=-\sum_{i=1}^{K}b_{i}\log{p_{b}(b_{i})}
\end{equation}
The total loss function can be written as:
\begin{equation}
l=l_{1}+\lambda_{i}l_{2}
\end{equation}
where $\lambda_i$ is the coefficient to balance two loss functions at i-th epoch. Since the bag-of-words generation module is built on the top of the word generation, we assign a small weight for the bag-of-words training at the initial time, and gradually increase the weight until a certain value $\lambda$:
\begin{equation}
\lambda_{i}=min(\lambda, k + \alpha i)
\end{equation}
In our experiments, we set the $\lambda=1.0$, $k=0.1$, and $\alpha=0.1$, based on the performance on the validation set.

\section{Experiments}

\begin{table*}[ht]
\centering
    \begin{tabular}{l|c|ccccc|c}
    \hline
    Model  & MT-02 & MT-03 & MT-04 & MT-05 & MT-06 & MT-08 & All\\ 
    \hline\hline
    Moses \citep{lattice}& 33.19 & 32.43 & 34.14 & 31.47 & 30.81 & 23.85 & 31.04        \\ 
    RNNSearch \citep{lattice}& 34.68 & 33.08 & 35.32 & 31.42 & 31.61 & 23.58 & 31.76              \\ 
    Lattice \citep{lattice} & 35.94  &  34.32 &  36.50 &  32.40 & 32.77  &  24.84 &  32.95  \\
    CPR \citep{PKI} & 33.84 & 31.18 & 33.26 & 30.67 & 29.63 & 22.38 & 29.72 \\
    POSTREG \citep{PKI} & 34.37 & 31.42 & 34.18 & 30.99 & 29.90 & 22.87 & 30.20 \\
    PKI \citep{PKI} & 36.10 & 33.64 & 36.48 & 33.08 & 32.90 & 24.63 & 32.51 \\
    Bi-Tree-LSTM \citep{bitreelstm} & 36.57 & 35.64 & 36.63 & 34.35 & 30.57 & - & -\\
    Mixed RNN \citep{mixedrnn} & 37.70 & 34.90 & 38.60 & 35.50 & 35.60 & - & - \\
     \hline\hline
     Seq2Seq+Attn (our implementation)&  34.71 & 33.15  & 35.26 & 32.36  & 32.45 &  23.96 & 31.96 \\
    \textbf{+Bag-of-Words (this paper)} &  \textbf{39.77} & \textbf{38.91}  & \textbf{40.02} & \textbf{36.82}  &  \textbf{35.93} &  \textbf{27.61} & \textbf{36.51} \\
    \hline
    \end{tabular}
    \caption{Results of our model and the baselines (directly reported in the referred articles) on the Chinese-English translation. ``-'' means that the studies did not test the models on the corresponding datasets.}
    \label{cnen}
\end{table*}

This section introduces the details of our experiments, including datasets, setups, baseline models as well as results.

\subsection{Datasets}
We evaluated our proposed model on the NIST translation task for Chinese-English translation and provided the analysis on the same task. We trained our model on 1.25M sentence pairs extracted from LDC corpora~\footnote{The corpora include LDC2002E18, LDC2003E07, LDC2003E14, Hansards portion of LDC2004T07, LDC2004T08 and LDC2005T06.}, with 27.9M Chinese words and 34.5M English words. We validated our model on the dataset for the NIST 2002 translation task and tested our model on that for the NIST 2003, 2004, 2005, 2006, 2008 translation tasks. We used the most frequent 50,000 words for both the Chinese vocabulary and the English vocabulary. The evaluation metric is BLEU \citep{bleu}.

\subsection{Setting}
We implement the models using \emph{PyTorch}, and the experiments are conducted on an \emph{NVIDIA 1080Ti} GPU. Both the size of word embedding and hidden size are 512, and the batch size is 64. We use Adam optimizer \citep{KingmaBa2014} to train the model with the default setting $\beta_{1}=0.9$, $\beta_{2}=0.999$ and $\epsilon=1\times10^{-8}$, and we initialize the learning rate to $0.0003$. 

Based on the performance on the development sets, we use a 3-layer LSTM as the encoder and a 2-layer LSTM as the decoder. We clip the gradients~\citep{gradientclip} to the maximum norm of 10.0. Dropout is used with the dropout rate set to 0.2. 
Following \citet{multichannel}, we use beam search with a beam width of 10 to generate translation for the evaluation and test, and we normalize the log-likelihood scores by sentence length.

\subsection{Baselines}

We compare our model with several NMT systems, and the results are directly reported in their articles.
\begin{itemize}
\item \textbf{Moses} is an open source phrase-based translation system
with default configurations and a 4-gram language model trained on the training data for the target language.
\item \textbf{RNNSearch}~\citep{attention} is a bidirectional GRU based model with the attention mechanism.
The results of Moses, and RNNSearch come from~\citet{lattice}.
\item \textbf{Lattice}~\citep{lattice} is a Seq2Seq model which encodes the sentences with multiple tokenizations.
\item \textbf{Bi-Tree-LSTM}~\citep{bitreelstm} is a tree-structured model which models source-side syntax.
\item \textbf{Mixed RNN}~\citep{mixedrnn} extends RNNSearch with a mixed RNN as the encoder.
\item \textbf{CPR}~\citep{gnmt} extends RNNSearch with a coverage penalty.
\item \textbf{POSTREG}~\citep{ganchev} extends RNNSearch with posterior regularization with a constrained posterior set.
The results of CPR, and POSTREG come from~\citet{PKI}.
\item \textbf{PKI}~\citep{PKI} extends RNNSearch with posterior regularization to integrate prior knowledge.
\end{itemize}

\subsection{Results}

Table~\ref{cnen} shows the overall results of the systems on the Chinese-English translation task. We compare our model with our implementation of Seq2Seq+Attention model. For fair comparison, the experimental setting of Seq2Seq+Attention is the same as BAT, so that we can regard it as our proposed model removing the bag-of-words target. The results show that our model achieves the BLEU score of 36.51 on the total test sets, which outperforms the Seq2Seq baseline by the BLEU of 4.55.

In order to further evaluate the performance of our model, we compare our model with the recent NMT systems which have been evaluated on the same training set and the test sets as ours. Their results are directly reported in the referred articles. As shown in Table~\ref{cnen}, our model achieves high BLEU scores on all of the NIST Machine Translation test sets, which demonstrates the efficiency of our model.

We also give two translation examples of our model. As shown in Table~\ref{model_example}, The translations of Seq2Seq+Attn  omit some words, such as ``\emph{of}'', ``\emph{committee}'', and ``\emph{protection}'', and contain some redundant words, like ``\emph{human chromosome}'' and ``$<$\emph{unk}$>$''. Compared with Seq2Seq, the translations of our model is more informative and adequate, with a better coverage of the bag-of-words of the references.


\begin{CJK}{UTF8}{gbsn}
\begin{table}[t]
\centering
    \begin{tabular}{p{7.2cm}}
    \hline
    \textbf{Source:} 人类共有二十三对染色体。\\
    \hline
    \textbf{Reference:} Humans have a total of 23 pairs of chromosomes . \\
    \hline
    \textbf{Seq2Seq+Attn:} Humans have 23 pairs chromosomes in human chromosome .\\
    \hline
    \textbf{+Bag-of-Words:} There are 23 pairs of chromosomes in mankind .\\
    \hline
    \hline
    \textbf{Source:} 一名奥林匹克筹备委员会官员说:「这项倡议代表筹委会对环保的敏感性。」\\
    \hline
    \textbf{Reference:} An official from the olympics organization committee said : `` this proposal represents the committee 's sensitivity to environmental protection . ''
\\
    \hline
    \textbf{Seq2Seq+Attn:} An official of the olympic preparatory committee said : `` this proposal represents the $<$unk$>$  of environmental sensitivity . ''
\\
    \hline
    \textbf{+Bag-of-Words:} An official of the olympic preparatory committee said : `` this proposal represents the sensitivity of the preparatory committee on environmental protection . '' \\
    \hline
    \end{tabular}
    \caption{Two translation examples of our model, compared with the Seq2Seq+Attn baseline.}
    \label{model_example}
\end{table}
\end{CJK}

\section{Related Work}

The studies of encoder-decoder framework \citep{Kalchbrenner,ChoEA2014,seq2seq} for this task launched the Neural Machine Translation. To improve the focus on the information in the encoder, \citet{attention} proposed the attention mechanism, which greatly improved the performance of the Seq2Seq model on NMT. Most of the existing NMT systems are based on the Seq2Seq model and the attention mechanism. Some of them have variant architectures to capture more information from the inputs~\citep{lattice,multichannel,Copy}, and some improve the attention mechanism~\citep{stanfordattention,interactive,supervisedattention,mapattention,fengattention,doublyattention}, which also enhanced the performance of the NMT model.

There are also some effective neural networks other RNN. \citet{fairseq} turned the RNN-based model into CNN-based model, which greatly improves the computation speed. \citet{googleattention} only used attention mechanism to build the model and showed outstanding performance. Also, some researches incorporated external knowledge and also achieved obvious improvement \citep{mixedrnn,bitreelstm}.

There is also a study~\cite{ZhaoEA2017} shares a similar name with this work, i.e. bag-of-word loss, our work has significant difference with this study. First, the methods are very different. The previous work uses the bag-of-word to constraint the latent variable, and the latent variable is the output of the encoder. However, we use the bag-of-word to supervise the distribution of the generated words, which is the output of the decoder. Compared with the previous work, our method directly supervises the predicted distribution to improve the whole model, including the encoder, the decoder and the output layer. On the contrary, the previous work only supervises the output of the encoder, and only the encoder is trained. Second, the motivations are quite different. The bag-of-word loss in the previous work is an assistant component, while the bag of word in this paper is a direct target. For example, in the paper you mentioned, the bag-of-word loss is a component of variational autoencoder to tackle the vanishing latent variable problem. In our paper, the bag of word is the representation of the unseen correct translations to tackle the data sparseness problem.


\section{Conclusions and Future Work}

We propose a method that regard both the reference translation (appears in the training set) and the bag-of-words as the targets of Seq2Seq at the training stage. Experimental results show that our model obtains better performance than the strong baseline models on a popular Chinese-English translation dataset. In the future, we will explore how to apply our method to other language pairs, especially the morphologically richer languages than English, and the low-resources languages.

\section*{Acknowledgements}

This work was supported in part by National Natural Science Foundation of China (No. 61673028), National High Technology Research and Development Program of China (863 Program, No. 2015AA015404), and the National Thousand Young Talents Program. Xu Sun is the corresponding author of this paper.

\nocite{SunEA2017,SunWei2017,dnerre,amr,discourse,wean,unpair,LinEA2018}

\bibliography{acl2018}
\bibliographystyle{acl_natbib}

\end{document}